# 3D Shape Classification Using Collaborative Representation based Projections

By  F. Fotopoulou[1], S. Oikonomou[2], A. Papathanasiou[2], G. Economou[2]
and S. Fotopoulos[2]

[1]Department of Computer Engineering and Informatics, University of Patras, 26500, Greece -  Email: fotopoulou@ceid.upatras.gr
[2]Department of Physics, University of Patras, 26500, Greece

**Abstract**
A novel 3D shape classification scheme, based on collaborative representation learning, is investigated in this work. A data-driven feature-extraction procedure, taking the form of a simple projection operator, is in the core of our methodology. Provided a shape database, a graph encapsulating the structural relationships among all the available shapes, is first constructed and then employed in defining low-dimensional sparse projections. The recently introduced method of CRPs (collaborative representation based projections), which is based on L2-Graph, is the first variant that is included towards this end. A second algorithm, that particularizes the CRPs to shape descriptors that are inherently nonnegative, is also introduced as potential alternative. In both cases, the weights in the graph reflecting the database structure are calculated so as to approximate each shape as a sparse linear combination of the remaining dataset objects. By way of solving a generalized eigenanalysis problem, a linear matrix operator is designed that will act as the feature extractor.
Two popular, inherently high dimensional descriptors, namely ShapeDNA and Global Point Signature (GPS), are employed in our experimentations with SHREC10, SHREC11 and SCHREC 15 datasets, where shape recognition is cast as a multi-class classification problem that is tackled by means of an SVM (support vector machine) acting within the reduced dimensional space of the crafted projections. The results are very promising and outperform state of the art methods, providing evidence about the highly discriminative nature of the introduced 3D shape representations.

1. **Introduction.**

The 3D shape classification as well as relevant tasks for shape recognition and retrieval remain highly challenging and demanding. Despite the fact that a plethora of methods has already been presented, there is still space for improvement. A critical point in all these tasks is the selection of a suitable descriptor for the reliable representation of 3D shape.

Regarding Global descriptors, spectral methods have dominated the field due to their isometry invariance property that makes them robust to deformations. ShapeDNA was one of the first descriptors obtained from the Laplace Beltrami

Operator (LBO) acting on the surface of the 3D shape [1], [23]. Other descriptors have also emerged from LBO with the GPS embedding one of the most well known [2]. This emerged from the Global Point Signature descriptor [20], with the important simplification of keeping only the eigenvalues, while ignoring the eigenfunctions. Biharmonic distance is also a global shape descriptor produced from LBO eigenvalues and proven very appropriate for classification [3],[4].

Recently a graph theoretic method was presented with very good shape classification performance [5]. The method enhances the sparse modeling objective function with the inclusion of a Laplacian term which preserves locality of encoded features during the sparse coding process. Beyond the classical machine learning methods, deep learning techniques have also presented [6].

In this work, we focus on a particular feature extraction scheme that is based on collaborative representation based projections (CRP). It is a novel dimensionality reduction technique [7] that results in a projection matrix after manipulating the sparse graph reflecting the estimated associations among the items of a given database. The term 'collaborative' points to the strategy adopted for representing each database item as a sparse combination of the rest items in the database. Briefly the method is as follows:

Given a set $X=\{X_1, X_2, X_3,…X_N\}$ of N m-dimensional (training set of) patterns, treated as column-vectors, the row-vector of reconstruction weights $W_i$ associated with the representation of $X_i$ when using as lexicon all the rest patterns is obtained by the following objective function based on L2 norm

$$W_i = \arg\min\{\|X_i - XW_i\|_2^2 + \lambda \|W_i\|^2\} \quad (1)$$

The N components of vector $W_i$ are the edge weights to be used for constructing the graph reflecting the structure of the data manifold. This is the main part of CRP method [7] and we denote this method hereafter as *L2Graph-CRP*. The sparsity constraint of (1) is much weaker than that of L1 graph [9], [19] but keeps the problem computationally tractable. In order to overcome this weak point, and based on L2 graph, a minimization procedure of local compactness has been included in the above framework [7]. This together with a maximization of total separability results in a highly discriminative projection matrix P that is capable of mediating data-driven feature extraction appropriate for classification tasks. Data-structure descriptions based on L1 graph, which has a very high computational cost, has also been successfully used for devising projections of similar nature (i.e. collaborative sparse representations) [9].

The objective of our work is twofold. First we adapt and apply CRP method to 3D shape classification and second we make an important modification by incorporating an algorithmic step that makes the data-structure graph representation better suited to the non-negative nature of the conventional shape-descriptors. Actually this is the case here for the ShapeDNA and GPS which take only nonnegative values and this constraint is expected to further enhance the classification performance of CPR method. The reconstructing problem is treated just as a quadratic optimization problem following the approximation method presented in [8] which is a Nonnegative Least Squares (NNLS) algorithm based on previous work on the Kuhn–Tucker theorem [10],[11].

Given a set $X=\{X_1, X_2, X_3,…X_N\}$ of non-negative patterns, every $X_i$ is reconstructed by minimizing:

$$\min_{W_i} \|XW_i - X_i\|_2^2 \quad \text{subject to } W_i \geq 0 \quad (2)$$

after excluding the $X_i$ itself from the patterns participating in its reconstruction.

In their well known text, Lawson and Hanson [8] gave the standard algorithm for NNLS - an active set method [22]. Mathworks modified the algorithm and gave the name *lsqnonneg*. The optimal number of the nonzero coefficients is indirectly defined and therefore the absence of sparsity constraint like that of (1) makes the reconstruction process direct and simpler. For our convention we name this method as *NNLS-CRP*.

Next, in part 2, the global shape descriptors will be presented, while the several steps of our framework will be detailed in part 3. In parts 4 and 5 experimental results, evaluation and conclusions will be given.

## 2. 3D shape representation and description

The performance of the method is greatly affected by the shape descriptor. We focus on the shape representation with global descriptors. Spectral based descriptors originate from the eigen-decomposition of Laplace-Beltrami Operator (LBO) applied to 3D shapes' surfaces S [20].

$$\Delta\Phi(s) = \lambda\Phi(s) \quad (3)$$

where $\Delta$ denotes the Laplacian operator defined over the geometry of S, and $\Phi(s)$ represents an eigenfunction with eigenvalue $\lambda$.

The eigenfunctions $\Phi_o, \Phi_1, \Phi_2$ … are placed in ascending order according to their eigenvalues $\lambda_0, \lambda_1, \lambda_2, ….$ It should be noticed that in the case of a closed manifold without boundary the first eigenvalue $\lambda_0$ is equal to zero and the $\Phi_o$ eigenvector is constant over the surface. Therefore does not contribute to the shape representation and is not included in the descriptors.

The truncated ordered sequence of the eigenvalues is the first spectral descriptor tested in this work with positive components named ShapeDNA [1].

$$\text{ShapeDNA} = \{\lambda_1, \lambda_2, \lambda_3, ...\} \quad (4)$$

The Global Point Signature (GPS) is another spectral descriptor developed from the inverse square roots of the LBO eigenvalues [2].

$$\text{GPS} = \{\frac{1}{\sqrt{\lambda_1}}, \frac{1}{\sqrt{\lambda_2}}, \frac{1}{\sqrt{\lambda_3}}, ....,\} \quad (5)$$

Both representations contain all the shape information for closed surfaces with no holes or handles [12] which is the usual case for most of the datasets, including SCREC11 which is of our interest here.

Both descriptors are isometry invariant and are proven to represent the 3D shapes very effective for retrieval, recognition, classification and clustering tasks. Spectral descriptors have dominated the representation of 3D shapes, however alternatives descriptors based on distributions of geodesic distances on the shape surface can be found in literature [21]. All of them are histograms of the distribution of the geodesic distances and therefore are positive-defined.

There are other variants based on LBO with improved properties. A brief list of global spectral descriptors is given in [13], [14]. It is not the objective of this work to compare the performance of various descriptors therefore we restrict ourselves to the above two descriptors: ShapeDNA and GPS. In both cases care is taken for scale invariance normalization. It is well known that scaling an $n$-dimensional manifold by a factor $a$ results in eigenvalues scaled by a factor $1/a^2$. Thus, by normalizing the eigenvalues, different shapes can be compared regardless of the object's scale and position [1]. In this work normalization is applied by simply dividing each descriptor (vector) by its magnitude

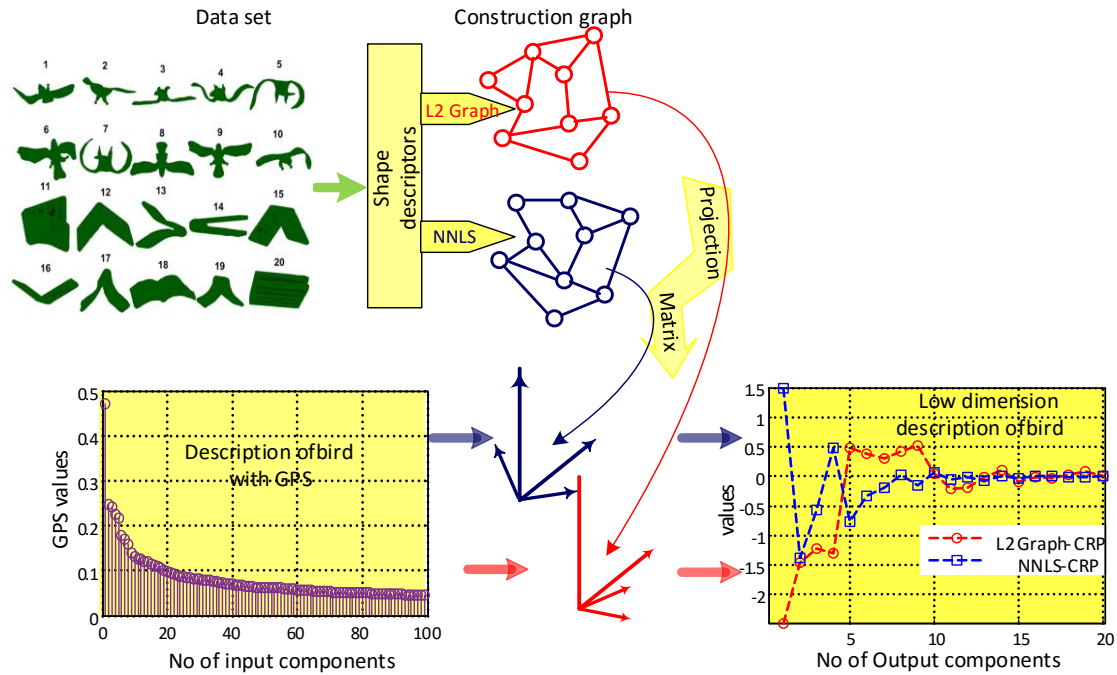

*Figure 1. The 3D shape ("bird") initially represented via GPS descriptor (shown in bottom-left figurine), which is of dimension 100, is "filtered" through a projection matrix and brought within a low dimensional space. The projection matrix has been derived based on a given shape dataset and conforms with the NNLS-CRP or L2Graph-CRP shape encoding. The nonzero values (seen in bottom-right figurine) are far less than the input dimension.*

The above 3D shape descriptors, usually, have a high dimension (>>100) in order to represent the shape precisely and convey all the information for an accurate classification. The objective of this work is to produce a parsimonious representation of the shapes. This goal is achieved by exploiting the structure description of a training dataset of shapes in order to build an optimal projection matrix. Without going to details at this section we give an initial example in figure 1 to highlight the issue of dimensionality reduction (final description) of the shape signatures. The input signature (GPS) with 100 nonzero values is reduced to less than 15 nonzero values. This final description of a shape is expected to offer an improved classification performance.

### 3. Our classification approach

The various stages of our method are visualized in the block diagram of Figure 2. Given the ensemble of 3D descriptors of a shape dataset, the linear mixture model finds the reconstructing weights to be used in the next step for deriving the graph, encompassing the dataset structure. The projection matrix is then computed providing high discrimination in the resulting low dimensional representation of the dataset. In the final step an SVM (Support Vector Machine) classifier is employed, acting on the projection coordinates to output the final estimation of the class label. The SVM parameters are set in the training phase where the label of every shape is known. The graph construction and the corresponding projection matrix are not explicitly trained, in the sense that the class label is not considered in this step.

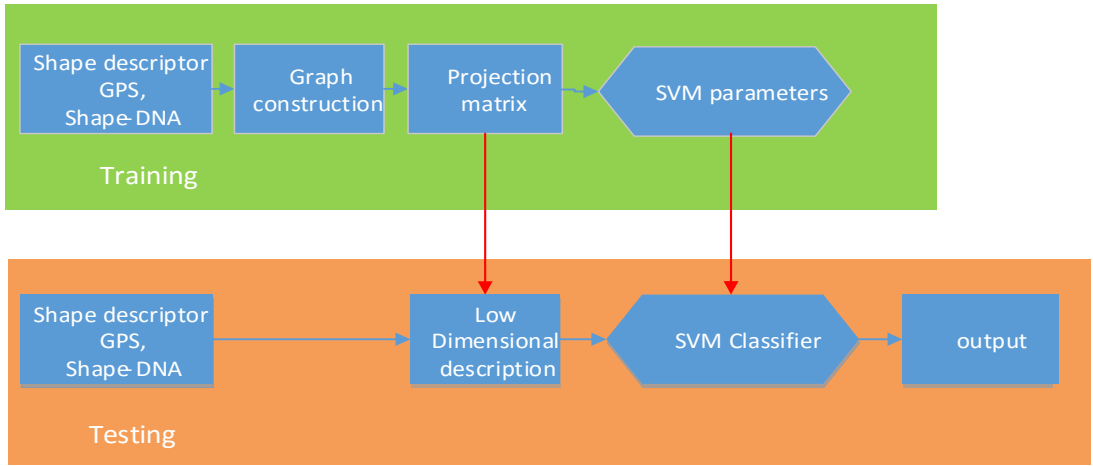

*Figure 2. Block diagram of our framework. In the training phase the projection matrix as well as the SVM parameters are defined. In the test phase the new parameters for each sample are provided by the projection matrix and are forwarded to the SVM classifier for class prediction.*

### 3.1 Graph construction – The nonnegative case

Against the standard graph construction methods where the k-nearest neighbors or the e-ball neighbors are used for node connections, we find the edge weights by reconstruction of each sample based on the rest of the set samples. Applying sparse representation techniques this idea has been successfully used for the construction of L2 graph as part of the CRP framework detailed in [7]. In a similar way we proceed in this part by describing NNLS-CRP, i.e the nonnegative version of the graph construction, where 3D shapes are represented with nonnegative descriptors. The reconstruction with nonnegative values is natural and meaningful and it is expected to have a positive impact on classification tasks. We briefly discuss the method and the tools associated for the specific task of graph creation when nonnegative samples are of interest. Every data point is approximated by an optimized linear mixture of the rest dataset items, using a nonlinear constraint. More specifically, given a set of N, m-dimensional vectors $X_i$ (i.e shape descriptors), each one of $X_i$s is approximated by the rest of the data set as a linear mixture of the form:

$$w_{i1}X_1 + w_{i2}X_2 + \cdots w_{i,j-1}X_{i-1} + 0X_i + w_{i,j+1}X_{i+1} \cdots + w_{iN}X_N = \mathbf{XW_i} \quad (6)$$

where **X** is the matrix with $X_i$s the data description vectors and $\mathbf{W_i}$ the N dimensional column vector of coefficients $\mathbf{W_i} = \{w_{ij}\}$, i=1:N, j=1:N..

These coefficients are found by minimizing the residual error $\mathbf{e_i = XW_i - X_i}$ in the sense of least squares optimization techniques. In our approach nonnegative coefficients $w_{ij}$, i=1:N, j=1:N are assumed. It should be noticed that this constraint is not mandatory but is used here for reasons referred to previously, and dealing with positive valued shape descriptors $X_i$ where their weighting coefficients $W_i = \{w_{ij}\}$ should be positive for a meaningful reconstruction. As was described in section 1 (introduction) this Nonnegative Least Squares (NNLS) problem presented by the functional (2) is efficiently solved by the algorithm described in Lawson and Hanson landmark text (1974) [8] and implemented in MATLAB by the function named `lsqnonneg`. The resulting coefficients are set as the edge weights of the graph i.e the edge weights between each $X_i$ and the rest of the nodes.

This fitting procedure defines the optimal number of nonzero coefficients resulting in a sparse representation of vector $W_i$ which finally assigns weight values to the graph edges. In the example given in figure 3, for the two shown shapes only 6 and 9, respectively, out of 600 coefficients are nonzero. In other words, only ~1% of the available shapes are necessary for reconstructing the two shapes. This could be considered as general result associated with the nonnegative constraint imposed to Least Square problem. At this point we mention that for L2graph-CRP the situation regarding sparsity is totally different. Many coefficients have significant values and the resulting graph is dense.

Regarding correct neighborhood selection, we observe that the shapes #95, #105 are reconstructed mostly with shapes of the same classes (81-100, 101-120).

This process to assign weights in the graph is very well suited for the 3D shapes usually represented with signatures in high dimensional space where choosing the nearest neighbors is not certain and unstable. The weighting vector $W_i$ is indicative of the class and can be used directly for classification in the same way as the sparse representation method is formulated to detect the correct classes. However better discrimination is attained by utilizing the coefficients for an additional step in the framework which is the graph creation. Elaborating on this graph, a projection matrix is produced that is instrumental for providing the final shape description. The SVM classifier is adopted for the final step of class assignment.

It is implicitly assumed that with the graph creation and the corresponding matrix the same weights $w_{ij}$ that reconstruct the $X_i$ point in the input space should also reconstruct the same point in the projected output space. Computing $W=\{W_i\}$ $i=1\ldots N$ the graph creation is completed.

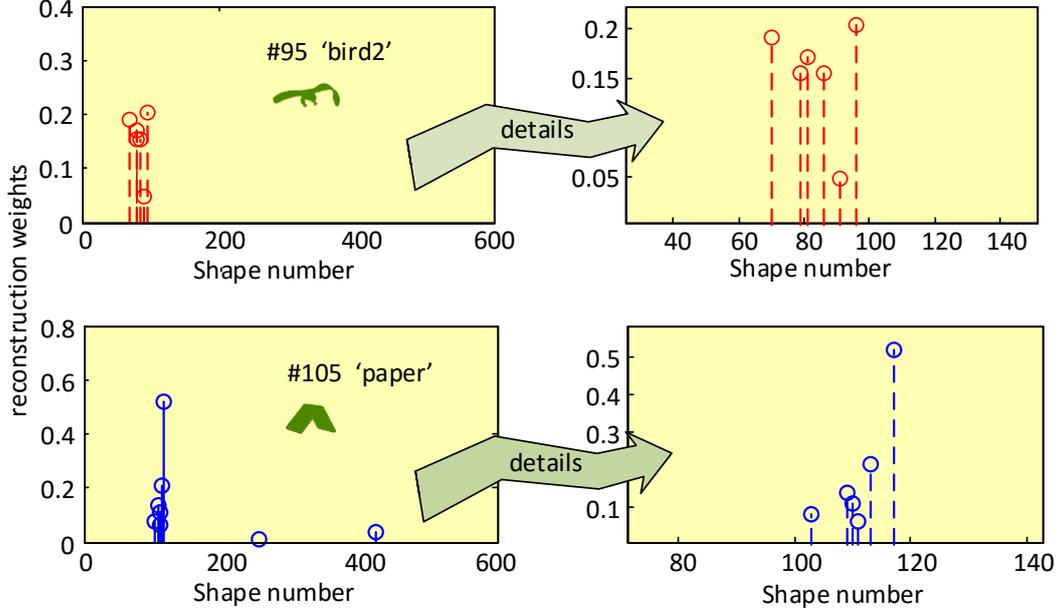

*Figure 3. Examples of the sets of reconstruction weights derived for two different shapes (#95 'bird2' and #105 'paper') using the encoding employed in NNLS-CRP method. The majority of reconstruction coefficients belongs to shapes of the same class. The sparsity of the coding coefficients is also evident.*

### 3.2 Projection matrix for NNLS-CRP

Having constructed the graph (X,W) we proceed to the next step to find the most discriminative projection matrix **P** based on the local compactness minimization and separability maximization [15].

The local compactness $J_C$ is defined as follows:

$$J_C = \sum_{i=1}^{N}\left\|P^T X_i - \sum_{i=1}^{N} W_{ij} P^T X_j\right\|^2 = P^T S_C P \quad (7)$$

where $S_C = X(I - W - W^T + WW^T)X^T$ is the local scatter matrix

The total separability $J_s$ is defined by

$$J_S = \sum_{i=1}^{N}\left\|P^T X_i - P^T \overline{X}\right\|^2 = P^T S_S P \quad (8)$$

where $\overline{X}$ is the average and $S_S = \sum_{i=1}^{N}(X_i - \overline{X})(X_i - \overline{X})^T$ is the scatter matrix of the whole data set.

The final function to be minimized is defined as follows:

$$J(P) = \arg\min_{P} \frac{P^T S_C P}{P^T S_S P} \quad (9)$$

and the desired projection matrix is computed by the largest eigenvectors of the equation

$$S_S P = \lambda S_C P \quad (10)$$

The projection matrix P finalizes the data description to feed the SVM classifier for the final class prediction. The new signature produced by the projection matrix is very descriptive of class identity due to the optimization of class separability (9). A visualization of this optimized discriminative property is given in figure 4. The distance between shapes computed as a component wise squared difference are given (a) between shapes of the same class as well as between shapes of different classes (b). The "within" class distance is less than 1% of the "between" class distance. The two classes randomly selected from SHREC2011 are numbered as 81-100 ('bird2') and 101-120 ('paper'). It should be noticed also that the contribution to distance is mainly due to the low dimension components, therefore we do not expect getting better by increasing the dimensionality of the projections. This remark is justified in section 4.2 (figure 6) where the impact of output dimension to classification accuracy is reported.

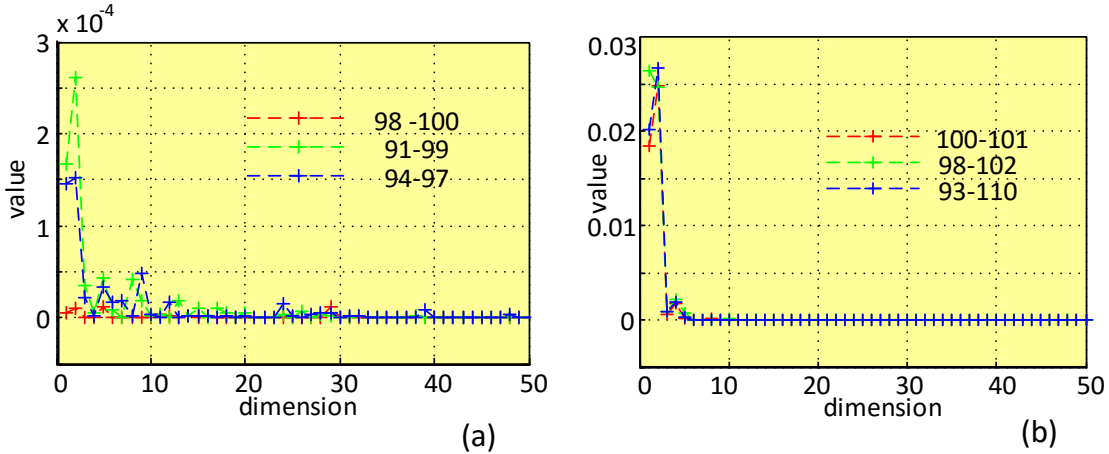

*Figure 4. Squared Difference along projected dimension for NNLS-GPS signatures. The shapes with labels 91 to 100 belong to the class 'bird2' while shapes 101 to 110 belong to class 'paper'. The Squared difference between shapes of the same class (a) is much less than between different classes (b). These values is a strong evidence of the optimized class discrimination due to the graph projection stage.*

### 3.3 Multiclass SVM

Multiclass Support Vector Machines (SVM) solve classification problems very effectively and their "one versus all" version have been extensively used in a wide range of applications. Specifically for a K number of classes K binary SVM classifiers are trained giving positive result every time for one class e.g the $i_{th}$ class and negative for the rest K-1 classes. The overall assignment to one of the K classes is based on the largest value of the decision function for the K binary problems.

## 3.4 Overview of the 3D shape classification method

All the above described stages of our framework are summarized in the algorithm1 comprised of the following steps:

**Algorithm 1**

Given the 3D shape Dataset $M=\{M_1,…M_N\}$

*step1. Find $X=\{X_1, X_2, ….X_N\}$, where the p-dimensional vector $X_i$s describes the 3D shape $M_i$s*

*step2. Compute the coefficients $W=\{W_1, W_2,….W_N\}$ and construct the corresponding graph ( either L2Graph or Least squares optimization-`lsqnonneg`)*

*step3. Find the matrix $P=\{P_1, P_2, ….P_d\}$ where $P_i$ are the eigenvectors of $S_S^{-1}S_C$ corresponding to d largest eigenvalues.*

*step 4. For an input 3Dshape with description vector $X_j$ find the final projection coordinates in the d dimensional space: $X_{jfinal}= P^T*X_j$*

*step 5. From $X_{jfinal}$ and the trained SVM find the class label of Xj*

## 4 Experimental results

Summarizing the proposed classification methods derived from our framework and with the two shape descriptors, we end up in four distinct cases:

- *L2Graph-GPS,*
- *L2Graph-ShapeDNA,*
- *NNLS-GPS*
- *NNLS-ShapeDNA.*

*L2Graph* refers to the CRP (collaborative representation based projections) method [7] with L2 graph, while *NNLS* to the nonnegative least squared version introduced in this work. *GPS or ShapeDNA* specifies the original shape descriptor signature.

The classification performance of our framework is evaluated through the above four instances of experimentation with SHREC-2010, SHREC-2011 and SHREC-2015, 3D shape benchmarks. SHREC-2011 nonrigid dataset [16],[17] , consists of 600 watertight triangle meshes i.e. 30 classes with 20 shapes in each class. Each class is labeled with 20 consecutive numbers. e.g numbers 81-100 stand for the class 'bird2' and 101 to 120 class 'paper'. In this work we refer several times in SHREC dataset to exemplify stages of our framework. SHREC-2010 dataset includes 200 mesh-type shapes from 10 classes [24]. Each class contains 20 shapes visually categorized in four sets : ants, crabs, spiders and octopi.

SHREC-2015 is a recent non-rigid 3D shape database consisting of 1200 watertight 3D triangle meshes equally divided into 50 classes with 24 shapes each [25].

Choosing these datasets allows us to compare against other recent state-of-the-art methods [5], [6], [26], [27], [28], [29], [30] and [31]. These results are reported in Section 4.3. In figure 5 a sample of 20 shapes from SHREC2011 are given selected from class 'bird2' and 'paper'. Classification accuracy (11) is the measure adopted here for an overall evaluation of our framework.

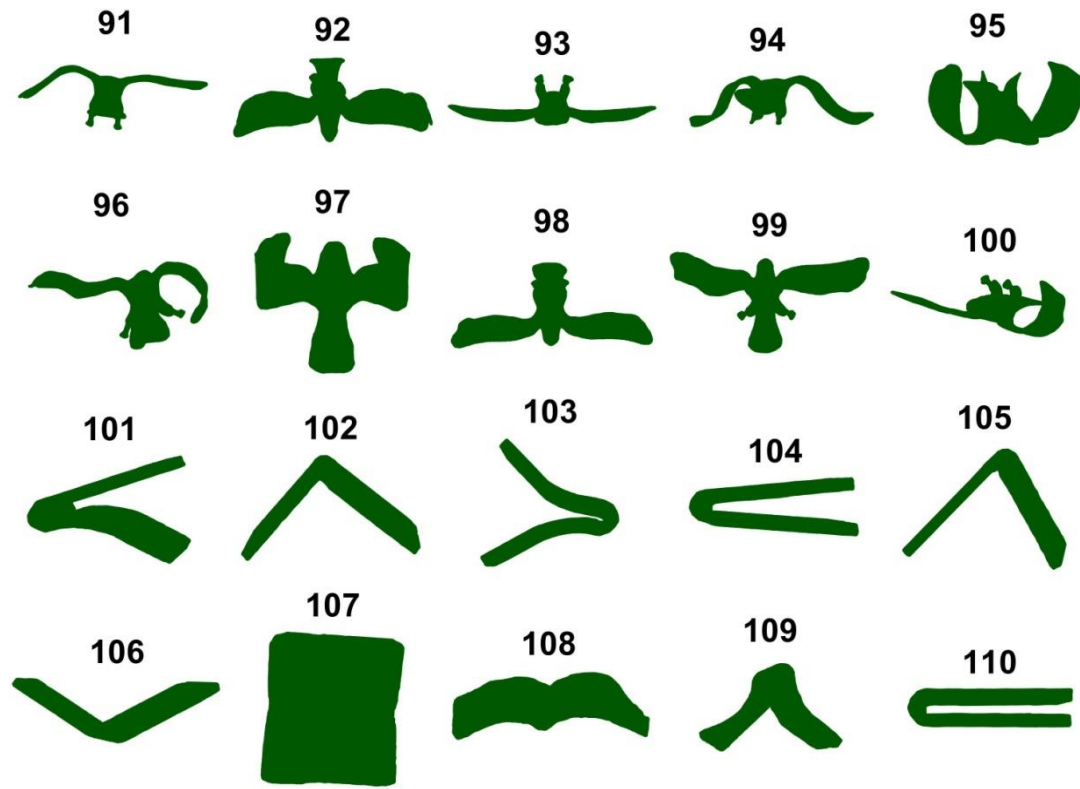

*Figure 5. A sample of 20 shapes from SHREC-2011 database . Shapes with labels 91 to 100 are from the class 'bird2' and 101 to 110 from class 'paper'. SHREC-2011 consists of 600 shapes and 20 consecutive numbers identify the members of each class.*

Classification accuracy could be derived from the confusion matrix. The main diagonal elements of this matrix contain the cases where the class labels given in the classification and ground truth data set agree. The sum of these elements i.e the correct classifications, divided by the total number of test instances provides a measure of classification accuracy. A direct definition is as follows:

$$accuracy = \frac{\# \, of \, correct \, classifications}{total \, \# \, of \, cases} \quad (11)$$

We followed three protocols for evaluation in order to make comparisons with various methods. In the first one select randomly 70% of shapes from the dataset in the training phase and keep the rest 30% for test. In the second we follow the 10 fold validation where the 90% is used for training and the rest 10% for test while in the third 50% of dataset is used for training and the rest 50% is kept for test. Training refers to graph construction as well as to learn SVM 'one vs all' model classifier. We apply the model to the test data and compute accuracy. This process is repeated 100 times and the average is taken as the overall classification accuracy.

## 4.1 Baseline Classification methods

In order to demonstrate the optimized performance of our method and based on SHREC2011 dataset, we present experimental results when the original descriptors (ShapeDNA [1], GPS [2]) are fed directly to an SVM classifier. Best performance is found by elaborating in the type and parameters of the adopted kernel. The dimension i.e the number of eigenvalues of the LBO (Laplace Beltrami operator) used in the shape signature was set to 10 following [5]. However no substantial change is reported by changing dimension.

The average classification accuracy for GPS-SVM was found ~94% and for the ShapeDNA-SVM ~93% . These results are included in the following Tables I to V.

Having computed this basic line performance our method will be assessed compared with the improvement of the above results.

## 4.2 Parameter settings

*Input shape signature – GPS, ShapeDNA dimension*

The truncated set of the ordered LBO eigenvalues describing the 3D shape define the input dimension. We set this dimension to 100. It is experimentally verified that in the present framework this dimension has a minor effect in the classification performance. For the L2graph-CRP the need for an over-complete dictionary which is a necessary constraint is always fulfilled in our experiments. The parameter $\lambda$ in eq.(1) which influences sparsity i.e the local compactness is set the typical value $\lambda=0.001N/700$ (with N=600) following the recommendation of [7].

*Optimum projection*

The projection matrix which is in the core of our framework provides the final components of the shape descriptor and their number $d$ should be set properly. Experimentation in SHREC2011 dataset, with increasing $d$ indicates that accuracy increases with dimension (d) reaching a wide region with small variation. Maximum accuracy is found for a certain dimension in this region. This observation is in line with results reported in section 3.2 based on Figure 4, where it was shown that the lower dimension components contribute mostly to shape discrimination. This is observed in both shape signatures: shapeDNA and GPS. In Figure 6 the accuracy vs (output) dimension is given for the SVM parameters with optimum classification performance for the L2Graph-ShapeDNA and NNLS-Shape DNA cases. Similar results are received for the GPS signature.

In the final step of our frame work is the SVM classifier. The linear c-SVM classifier is realized in all cases.

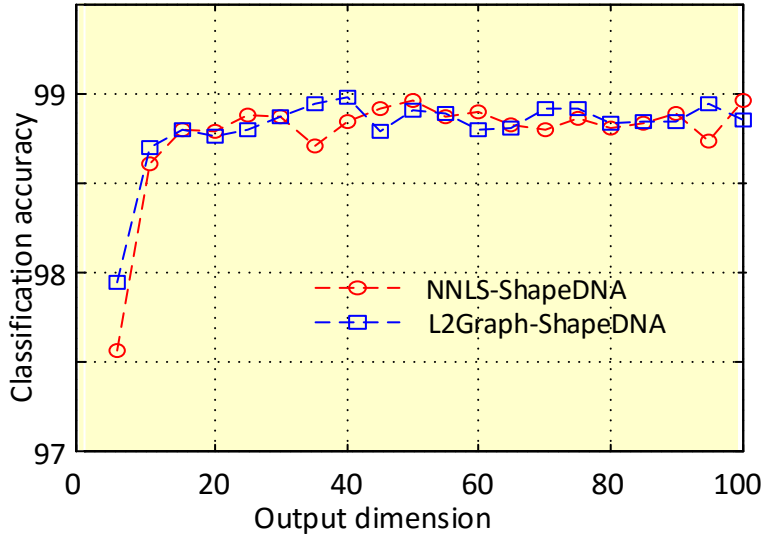

*Figure 6. Accuracy vs output dimension for L2Graph-ShapeDNA, and NNLS-Shape DNA. The linear c-SVM classifier is implemented in both cases and tuned to the optimum performance. The training set was 70% of the given dataset (SHREC-2011).*

### 4.3 Classification results

*SHREC2011.* It is the largest non-rigid 3D model dataset which is publicly available and it is extensively used in retrieval and classification experimentation.

Results of the accuracy are summarized in TABLE I together with a few basic and state of the art methods. ShapeDNA-SVM and GPS-SVM indicate the basic line methods where the shape descriptors are fed directly to SVM classifier. GraphBDM is the classification method proposed in [5] and DeepGM in [26] with given accuracy for the same data set (SHREC2011) and training with 70% of the dataset objects. GWCNN is proposed in [31] using for training 80% of the dataset. GWCNN is a Deep neural network technique. Results reported in [28], [29] and [30] are also included. Their placement in the second column of the Table I is arbitrary. To our knowledge accuracy results of [5],[26] are the highest among similar methods The different classification performance of GPS, ShapeDNA, emphasizes the impact of descriptor in the discriminative representation of shapes as well as the algorithms employed in the framework.

| TABLE I SHREC2011 Classification Accuracy |||
| --- | --- | --- |
| Method | Training with 70% of the data set | Training with 90% of the data set |
| ShapeDNA-SVM | 92.76 | 93.07 |
| GPS-SVM | 93.61 | 94.20 |
| Method of [30] | - | 96.00 |
| GWCNN [31] | 96.6 | - |
| Graph BDM [5] | 97.59 | - |
| Deep GM [26] | 97.89 | - |
| CTA [28] | 97.0 | - |
| NNLS-ShapeDNA | 98.95 | 98.70 |
| NNLS-GPS | 98.61 | 98.95 |
| 3DVFF [29] | - | 99.1 |
| L2Graph-ShapeDNA | 98.97 | 99.13 |
| L2Graph-GPS | **99.10** | **99.30** |

*SHREC2010.* Results are in Table II. L2Graph-GPS shows the best scores in both training setups. The size of the training set has a serious impact on the accuracy especially in the NNLS method. This could be attributed to the small size of the SHREC2011 dataset. Also ShapeDNA has a poor performance in this dataset.

| TABLE II | | |
|---|---|---|
| **SHREC2010** Classification accuracy | | |
| Method | Training with 50% of the data set | Training with 70% of the data set |
| L2Graph-GPS | **97.56** | **98.75** |
| NNLS-GPS | 89.92 | 95.80 |
| L2Graph-ShapeDNA | 75.5 | 78.7 |
| NNLS-Shape DNA | 75.2 | 78.25 |
| DeepSGW [6] | 96.00 | - |
| SGWC-BoF [27] | 95.66 | - |
| GraphBDM [5] | - | 96.67 |

*SHREC2015.* Results are given in Table III for 2 of our methods with best performance: L2Graph-GPS and NNLS-GPS. DeepGM [26] which outperforms our method is based to recently presented Deep Learning techniques. It should be noticed that *SHREC2015* dataset includes models with 2 or 3 zero eigenvalues that are removed weakening the discriminative power of GPS descriptor. It is also noted that in this case of experimentation each shape of SHREC2015 is represented with a feature vector of dimension 200.

| TABLE III | |
|---|---|
| **SHREC2015** Classification accuracy | |
| Method | Training with 70% of the data set |
| L2Graph-GPS | 91.84 |
| NNLS-GPS | 91.28 |
| L2Graph-ShapeDNA | 87.94 |
| NNLS-Shape DNA | 87.56 |
| DeepGM [26] | 93.03 |

These classification results presented in TABLES I to III for the two training conditions, indicate that:

- All four implementations do not differ significantly in respect to classification accuracy. This result proves that the NNLS-CRP is a reliable alternative to CRP (collaborative graph embedding projection) method which is based on L2Graph construction.
- Among the four presented methods the L2Graph-GPS has the best performance.
- Compared to similar state of the art methods where data are available, all four methods have better performance in SHREC2011 dataset, best performance of L2Graph-GPS in SHREC2010, and comparable results in SHREC2015 dataset
- In all cases the inclusion of graph projection stage improves the classification accuracy. This improvement which is more than 5% refers to the baseline method i.e when the GPS (or ShapeDNA) descriptor feeds directly the SVM classifier.

## 5      Conclusions

In this work we presented a framework where a collaborative representation based projections (CRP) technique [7] is used in conjunction with an SVM classifier for the purpose of 3D shape classification. The projection matrix is based on a graph creation technique using either an L2graph [18] or a NNLS (Nonnegative constrained Least Square) technique [8]. Inter-class separability is enhanced by means of a projection that minimizes local compactness to total separability [7],[15]. The nonnegative case, NNLS-CRP, was introduced as a fast and simpler alternative to L2Graph-CRP expecting to work efficiently with nonnegative descriptors (like ShapeDNA, GPS). Our objective was to demonstrate that the inclusion of the projection matrix in the framework enhances classification accuracy. In this regard the essential step is the graph construction providing stability to high dimensional descriptors especially when manually choosing the nearest neighbors fails due to high instability. It was beyond our scope to make a thorough study of different 3D shape descriptors regarding their impact on classification accuracy.  However using two types of well known global descriptors (i.e GPS and ShapeDNA) we demonstrated that our method reached state of the art performance and with fewer parameters than other competitive methods. Looking for a better shape descriptor is the obvious extension of the present work. The method could also be extended to other 3D shape classification, clustering and recognition tasks without substantial modifications.

An updated version of this paper is under consideration at Pattern Recognition Letters.